**Predicting Barge Tow Size on Inland Waterways Using Vessel Trajectory Derived Features: Proof of Concept**

**Geoffery Agorku***
Department of Civil Engineering
University of Arkansas, Fayetteville, Arkansas, 72701
gagorku@uark.edu
https://orcid.org/0009-0000-7716-0334

**Sarah Hernandez, PhD, P.E**
Department of Civil Engineering
University of Arkansas, Fayetteville, Arkansas, 72701
sarahvh@uark.edu
https://orcid.org/0000-0002-4243-1461

**Hayley Hames**
Department of Geosciences
University of Arkansas, Fayetteville, Arkansas, 72701
hhames@uark.edu
https://orcid.org/0009-0002-7285-3005

**Cade Wagner**
Department of Department of Mathematical Sciences
University of Arkansas, Fayetteville, Arkansas, 72701
cadew@uark.edu
https://orcid.org/0009-0000-1650-6450

Word Count: 6,381 words + 2 tables (250 words per table) = 6,881 words

Submitted *[10/27/2025]*
*Corresponding Author

## ABSTRACT

Accurate, real-time estimation of barge quantity on inland waterways remains a critical challenge due to the non-self-propelled nature of barges and the limitations of existing monitoring systems. This study introduces a novel method to use Automatic Identification System (AIS) vessel tracking data to predict the number of barges in tow using Machine Learning (ML). To train and test the model, barge instances were manually annotated from satellite scenes across the Lower Mississippi River. Labeled images were matched to AIS vessel tracks using a spatiotemporal matching procedure. A comprehensive set of 30 AIS-derived features capturing vessel geometry, dynamic movement, and trajectory patterns were created and evaluated using Recursive Feature Elimination (RFE) to identify the most predictive variables. Six regression models, including ensemble, kernel-based, and generalized linear approaches, were trained and evaluated. The Poisson Regressor model yielded the best performance, achieving a Mean Absolute Error (MAE) of 1.92 barges using 12 of the 30 features. The feature importance analysis revealed that metrics capturing vessel maneuverability such as course entropy, speed variability and trip length were most predictive of barge count. The proposed approach provides a scalable, readily implementable method for enhancing Maritime Domain Awareness (MDA), with strong potential applications in lock scheduling, port management, and freight planning. Future work will expand the proof of concept presented here to explore model transferability to other inland rivers with differing operational and environmental conditions.

## 1. INTRODUCTION

The U.S. inland waterway system, an extensive network of approximately 25,000 miles of navigable waters, is a vital artery for the nation's economy, moving over 600 million tons of cargo annually (*1*). Yet, the inability to systematically track the primary cargo-carrying unit (the barge) in real-time remains a challenge (*2*). This creates a critical blind spot that hampers efficiency, compromises safety, and limits strategic planning across the entire waterway ecosystem (*3*).

The core of the issue stems from the nature of barge transportation itself. Unlike the self-propelled tug and towboats that are mandated to carry Automatic Identification System (AIS) transponders, the barges they push are non-self-propelled, unpowered assets. The AIS system uses Very High Frequency (VHF) radio transponders to automatically broadcast vital information, such as a vessel's identity, position, course, and speed, to other ships and shore stations for navigation and collision avoidance (*4*). Barges, however, are the floating equivalent of a truck trailer or a railcar but without the standardized tracking mechanisms common to those modes. Consequently, while the location of the tug is generally known, the quantity, configuration, and status of the actual cargo units remain unknown. The result is a system where operational decisions are often based on incomplete, outdated, or inaccurate information (*2*).

Inland waterways possess inherent and significant advantages over road and rail for the transport of bulk commodities, particularly in terms of cost-effectiveness and environmental sustainability (*5*). However, a key barrier that has historically prevented the full realization of this potential is the mode's lack of visibility and its perceived unreliability, the "black box" problem (*6*). Shippers of higher-value or more time-sensitive goods have often opted for more expensive but more transparent modes like trucking or rail, effectively paying a premium for visibility and predictability (*7*). A robust, real-time barge tracking and counting system directly addresses this competitive disadvantage, unlocking profound benefits for commercial operators, public agencies, and the national supply chain (*8*).

The objectives of this paper are as follows:

i. To estimate the number of barges being towed by a tug/towboat with minimal error, leveraging both static vessel geometry and dynamic movement patterns derived from AIS data,
ii. To identify the most influential predictors of barge count, and
iii. To establish a foundation for future spatial generalization of barge count prediction models across diverse riverine environments.

## 2. BACKGROUND

Current barge monitoring systems are a patchwork of legacy methods limited by high latency, lack of granularity, and incomplete coverage. These systems fall short of meeting the demands of modern, real-time logistics (*2*). New technologies to address barge tracking range from inferential software models to direct physical observation (*2*, *9*). Each of these approaches offers a distinct set of strengths, weaknesses, and suitability for different operational needs (*2*, *10*). As an example, the Lock Performance Monitoring System (LPMS), records tonnage and volume but is inherently limited to the approximately 12,000 miles of the U.S. Marine Highway system served by locks. This leaves 17,000-miles, including stretches of free-flowing rivers, unmonitored by this system (*2*). Other national-level datasets, such as the Commodity Flow Survey (CFS) and the Waterborne Commerce Statistics Center (WCSC), provide macroeconomic perspectives but lack spatial disaggregation needed to track movements at the level of an individual waterway, let alone a specific barge tow (*11–14*). Furthermore, these systems suffer from structural limitations, including delays in data collection and publication, potential sampling biases, and issues with data quality (*13*, *14*). Traditional commercial methods, such as waybills, are similarly flawed. They rely on manual data entry, which is prone to human error, and are subject to reporting delays that are incompatible with the needs of an efficient, real-time monitoring system (*15*, *16*).

The challenges of barge monitoring on inland waterways are being addressed by a new generation of technologies. Four primary methodologies for determining real-time barge counts are referenced in the literature.

*Inferential Analysis using Machine Learning on AIS Data* (*17*)*:* Rather than directly observing barges, this technique infers their presence and quantity by analyzing the movement patterns and behavior of tugboats. Its primary strength lies in its scalability and cost-effectiveness. It leverages the widespread AIS infrastructure, which provides continuous data from nearly all commercial tugs operating on inland waterways. As a result, this method can be implemented without the need to install any new hardware. However, because it does not observe barges directly, the accuracy of barge counts is dependent on the reliability of inference models.

*Direct Observation using Computer Vision* (*2*, *18*, *19*)*:* This approach uses deep learning models applied to video feeds from existing cameras to visually detect and count barges. Its main advantage is direct visual confirmation, making it a dependable source of ground-truth data. Additionally, it can be deployed relatively inexpensively by utilizing existing public camera infrastructure. This method is limited by the field of view of available cameras and is vulnerable to environmental factors such as poor lighting, fog, rain, or obstructed viewing angles. Scaling this solution would require installing additional cameras at strategic locations or identifying more publicly accessible feeds that provide clear views of the waterways.

*Asset-Level Intelligence using Dedicated Internet of Things (IoT) Tracking Devices* (*20*, *21*)*:* This strategy involves placing satellite or cellular-based trackers directly on individual barges, turning them into connected, intelligent assets. This approach offers direct, real-time data on the location and status of each barge. It is valuable for tracking high-value or hazardous cargo. Satellite-based trackers also offer coverage in remote areas that lack landside infrastructure. However, the method carries significant costs in hardware and ongoing data subscriptions. It also requires logistical coordination to deploy and retrieve devices at the beginning and end of each voyage. Adoption of this approach depends heavily on the willingness of barge owners and operators to invest in the technology.

*Wide-Area Surveillance using Satellite and Aerial Imagery* (*22*–*27*)*:* This approach enables monitoring of vast geographic areas, including remote or infrastructure-poor segments of inland waterways. It is suited for detecting vessels operating without AIS ("dark vessels") or monitoring activity near sensitive sites. This method is limited in real-time operational contexts as satellite revisit times can range from hours to days. Many satellite sensors lack the resolution necessary to distinguish individual barges in tightly grouped tow configurations. The high costs associated with acquiring and processing such imagery further limit its feasibility for continuous operational use.

These four technologies can serve as complementary components of a multi-layered data ecosystem. A robust and resilient barge tracking network will likely be a hybrid "system of systems" that leverages the strengths of each approach to compensate for the weaknesses of others (*28*). A logical architecture for such a system would begin with the deployment of direct observation technologies, like computer vision from satellite imagery, which would serve as "ground truth", capturing high-fidelity, labeled data of actual tow configurations (*29*, *30*). This verified data can then be used for training and continuous validation for the highly scalable, low-cost inferential models that run on only AIS data. In this model, the computer vision model acts as the "teacher," constantly refining the accuracy of the machine learning barge prediction algorithm, which this paper implements.

The availability of precise, real-time data on barge counts and tow configurations has the potential to better manage the inland waterway system in terms of lock and dam operations, port and terminal efficiency, and informing planning and policy. The nation's lock and dam system is the primary source of bottlenecks on the inland waterways (*31*). 80% of waterway infrastructure has exceeded its 50-year design life, leading to frequent delays and closures (*32*). A primary driver of inefficiency at these locks is the mismatch between modern tow sizes and legacy lock dimensions (*33*). A standard lock chamber on many U.S. rivers is 600 feet long, while a typical modern tow can be twice that length. To transit, this large tow must perform a time-consuming "double lockage" (*34*). Currently, lock operators often have no advance knowledge of whether an approaching vessel will require a single or double lockage, forcing them into a reactive mode. The ability to predict the exact number of barges in an approaching tow in real-time would allow lock masters to optimize queuing, pre-position resources, and manage water levels. The same principle of proactive management applies to the operations of river ports and terminals. For these commercial hubs, ambiguity about arriving vessel configurations translates directly into wasted time, underutilized assets, and increased costs. Studies confirm that such operational procedures can decrease system-wide delays, and real-time data would enable these advanced traffic control strategies (*2, 35, 36*).

Real-time data provides value to public agencies like the U.S. Army Corps of Engineers (USACE), the Departments of Transportation (DOTs), and the Maritime Administration (MARAD), who are responsible for planning, performance monitoring, and economic analysis. This capability represents a transition from relying on latent, highly aggregated statistics to a dynamic, live view of the system. Planners can precisely measure traffic volumes, identify emerging bottlenecks, and analyze the impact of disruptions in real-time (*37*). This detailed knowledge is essential for understanding commodity flows and provides a powerful tool for long-range transportation planning (*38*). Furthermore, legislative funding programs like the Infrastructure Investment and Jobs Act (IIJA) require data-based justification for project investment. Accurate, real-time data on traffic volumes, tow sizes, and delays provides the empirical evidence needed to prioritize projects with the highest potential.

## 3.    METHODS

This section presents a framework for detecting, classifying, and analyzing vessel and barge activity using satellite and AIS data. The methodology integrates computer vision, spatiotemporal data fusion, trajectory analysis, and machine learning to estimate barge counts and vessel operational characteristics. It is designed to be transferable across different river systems or inland waterway contexts (**Figure 1**).

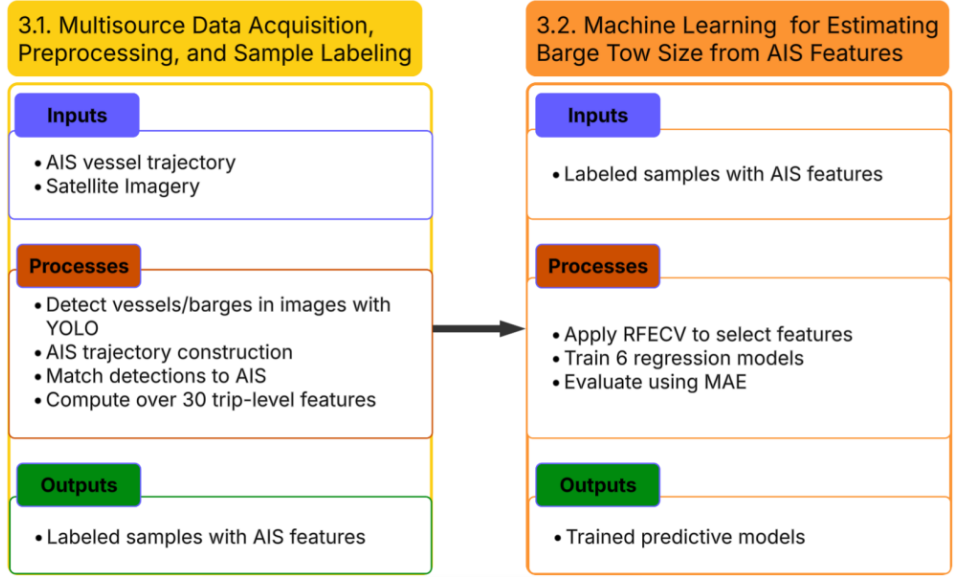

**Figure 1: Methodological Framework for Predicting Barge Counts from Vessel Trajectories**

### 3.1. Multisource Data Acquisition, Preprocessing, and Sample Labeling

The methodology outlined in this paper requires two complementary datasets for implementation.

- *Vessel trajectories* are obtained from AIS data, which include vessel IDs, locations, speeds, and directions. These data are accessible through both public and commercial providers.
- *Satellite imagery* is collected from either commercial sources (e.g., Planet Labs or Maxar) or public platforms (e.g., Sentinel-2). To maintain temporal consistency, satellite images are selected to match the timeframes when AIS data is available for the same geographic area.

Although the aim of this paper is to use only AIS derived features to predict barge count, an interim process to ease manual sample labeling is introduced. This process applies pre-trained machine learning models (e.g., You Only Look Once, YOLO (*39*)) to automatically detect vessel and barge objects from satellite imagery which are then later manually classified by two researchers and matched to AIS records. Then, to associate vessels detected in satellite imagery with those reported through AIS, a multi-step spatiotemporal integration procedure is employed. After satellite and AIS data are related through spatiotemporal matching, the AIS points are converted into trajectories defined by an origin and destination and features are extracted.

*Spatiotemporal Matching for Labeled Samples*
For every satellite scene, the exact image acquisition time is recorded. AIS messages transmitted within two minutes before or after this timestamp are selected, creating a narrow temporal window that reduces mismatches while allowing for minor differences between AIS reporting and image capture times. After vessels are identified in the imagery using a YOLO-based detector and their bounding boxes are georeferenced, these filtered AIS records provide candidate matches.

Next, the selected AIS records are grouped by vessel identifier (e.g., MMSI) and chronologically ordered to form paths representing vessel movement. These sequences of points are connected into continuous polylines, revealing the travel direction, speed, and spatial extent within the imaged region. Finally, each satellite-derived bounding box is converted into a geospatial polygon using metadata that links image pixels to geographic coordinates. A detection is considered a valid match if its polygon intersects with an AIS trajectory and falls within the defined temporal window. When a match is established, the associated AIS attributes such as vessel identity, position, heading, and velocity are transferred to the detection (**Figure 2**).

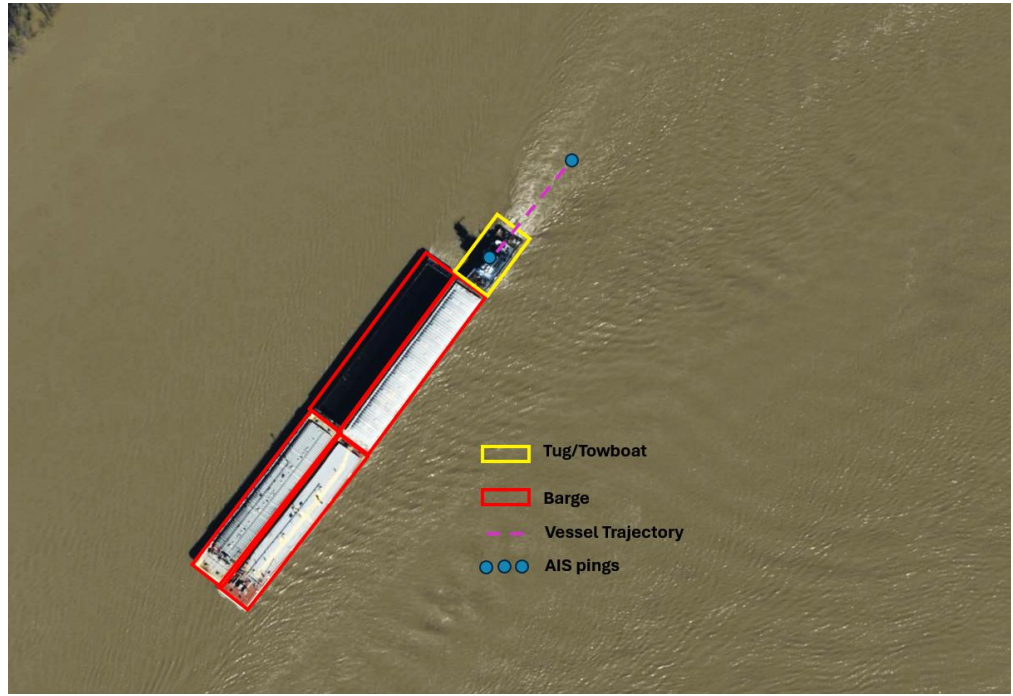

**Figure 2: Example of spatiotemporal fusion of satellite imagery and AIS data**

*AIS Trajectory Reconstruction*

Once AIS metadata is linked to the detected vessels, vessel trips are extracted by identifying stop locations using a density-based clustering algorithm, following the approach proposed by Asborno et al (*40*). This method incorporates low-speed thresholds (e.g., <2 knots), minimum stop durations, and the spatial clustering of AIS pings. These stops serve as origin and destination points, with each trip defined as the movement segment between two consecutive stop points. It is assumed that barge configurations remain unchanged throughout the duration of each trip and the resulting models for barge count prediction only apply to moving vessels.

To determine which specific trip corresponds to the satellite imagery, the satellite image's acquisition timestamp is compared against the sequence of trips predicted for the vessel. The trip that includes the satellite capture time (i.e., the one whose start time is immediately before and end time is immediately after the satellite timestamp) is selected as the matching trip. This allows for accurate attribution of barge configuration and movement characteristics during the time the satellite image was taken.

*AIS Feature Extraction*

AIS points from the matched trip were processed to compute over 30 trip-level statistics ("trip features") describing vessel geometry, kinematic behavior, and river context (**Table 1**). These features were grouped into categories such as speed and acceleration dynamics, directional stability, and operational patterns. The feature set was guided by the premise that vessels towing larger barge configurations tend to exhibit reduced maneuverability, smoother courses, and fewer directional changes supporting the hypothesis that motion-based features can serve as proxies for inferring barge counts.

**Table 1: Features extracted from AIS-derived vessel trajectories for barge count prediction models**

| Feature Group | Feature | Description | Formula | Acronym |
|---|---|---|---|---|
| **Vessel Size** | Vessel Length | Hull length in meters | L | LEN |
| | Vessel Width | Maximum hull beam width | W | WID |
| | Vessel Draft | Vessel draft depth | D | DFT |
| **Speed** | Mean Speed | Average vessel speed | $\bar{s} = \dfrac{1}{n}\sum_{i=1}^{n} s_i$ | SOG_MEAN |
| | Median Speed | Typical vessel speed | $s_{med} = median\,(S)$ | SOG_MED |
| | Standard Deviation of Speed | Speed variability measure | $\sigma_s = \sqrt{\dfrac{1}{n-1}\sum_{i=1}^{n}(s_i - \bar{s})^2}$ | SOG_STD |
| | Interquartile Range of Speed | Mid-range speed spread | $IQR = Q3_s - Q1_s$ | SOG_IQR |
| | Median Absolute Deviation of Speed | Robust speed dispersion | $MAD_s = median(|s_i - s_{med}|)$ | SOG_MAD |
| | Maximum Observed Speed | Peak observed speed | $S_{max} = max(S)$ | SOG_MAX |
| | Minimum Observed Speed | Lowest observed speed | $S_{min} = min(S)$ | SOG_MIN |
| | Speed Range | Total speed span | $S_{range} = S_{max} - S_{min}$ | SOG_RANGE |
| | Coefficient of Variation of Speed | Relative speed variability | $CV_s = \dfrac{\sigma_s}{\bar{S}}$ | SOG_CV |
| | Percent Time at Low Speed | Time at near-idle speed | | SOG_PCT_LOW |
| | Percent Time at High Speed | Time above high-speed | $p_{Cond} = \dfrac{Count\left(\begin{array}{c}s_i\ where\\ condition\ is\ true\end{array}\right)}{n} \times 100$ | SOG_PCT_HIGH |
| | Percent Time at Optimal Speed | Time within optimal range | | SOG_PCT_OPT |
| | Entropy of Speed Distribution | Speed distribution entropy | $Hs = \sum_{k} P(s_k)\,log_2\,P(s_k)$ | SOG_ENT |
| **Acceleration & Stability** | Mean Positive Acceleration | Average speed-up acceleration rate | $\bar{a}_{pos} = \dfrac{1}{|A^+|}\sum_{a \in A^+} a$ $[A^+ = \{a\ \epsilon A | a > 0\}]$ | ACC_POS_MEAN |
| | Mean Negative Acceleration | Average deceleration magnitude | $\bar{a}_{neg} = \dfrac{1}{|A^-|}\sum_{a \in A^-} a$ $[A^- = \{a\ \epsilon A | a > 0\}]$ | ACC_NEG_MEAN |
| | Standard Deviation of Acceleration | Acceleration fluctuation measure | $\sigma_a = \sqrt{\dfrac{1}{n-2}\sum_{i=1}^{n-1}(a_i - \bar{a})^2}$ | ACC_STD |
| | Maximum Deceleration | Maximum deceleration event | $a_{dec\_max} = min(A)$ | ACC_MIN |

| Feature Group | Feature | Description | Formula | Acronym |
|---|---|---|---|---|
| | Acceleration Sign Changes | Count of acceleration reversals | $ZC_a = \sum_{i=1}^{n-2} 1\,(sign(a_i) \neq (sign(a_{i+1}))$ <br> 1: indicator function | ACC_ZC |
| **Course & Heading Behavior** | Standard Deviation of Course | Course variability measure | $\sigma_c = \sqrt{\frac{1}{n-2}\sum_{i=1}^{n-1}(c_i - \bar{c})^2}$ | COG_STD |
| | Entropy of Course Over Ground | Course distribution entropy | $H(C) = \sum_k P(c_k)\,log_2\,P(c_k)$ | COG_ENT |
| | Standard Deviation of Turn Rate | Turn-rate fluctuation measure | $\sigma_{tr} = std\left(\lvert\frac{C_{i+1} - C_i}{\Delta t}\rvert\right)$ | TRN_STD |
| | Total Absolute Course Change | Cumulative course change | $\Delta C_{tot} = \sum_{i=1}^{n=1}\lvert C_{i+1} - C_i\rvert$ | COG_TOTAL_CHANGE |
| | Mean Difference between Course Over Ground and Heading | Mean heading offset | $\Delta_{CH} = \{c_1 - h_1 \ldots c_n - h_n\}$ <br> $\mu\Delta CH = mean(\Delta_{CH})$ | COG_HDG_DIFF_MEAN |
| | Standard Deviation of Course Over Ground and Heading Difference | Heading offset variability | $\sigma\Delta CH = std(\Delta_{CH})$ | COG_HDG_DIFF_STD |
| **Trip Efficiency & Geo** | Total Trip Time | Total voyage duration | $T_t = (n-1) \times \Delta t$ | DUR_HRS |
| | Distance Traveled | Total voyage path length | $D_{t=\sum_{i=1}^{n-1}\bar{S}_i\times\Delta t}$ <br> $\bar{S}_i = \frac{S_i + S_{i+1}}{2}$ | DIST_KM |
| | Direct Straight Line Distance | Straight-line displacement | $D_{direct} = Haversine\left(\begin{matrix}lat_1, lon_1, \\ lat_n, lon_n\end{matrix}\right)$ | DIST_HAVERSINE_KM |
| | Sinuosity Index | Path tortuosity index | $sino = \frac{D_{trip}}{D_{direct}}$ | SINO_IDX |
| **Interaction & Polynomials** | Vessel Area | Hull footprint area | $A_{vessel} = L \times W$ | AREA |
| | Draft to Length Ratio | Draft / Length | $R_{DL} = \frac{D}{L}$ | DLT_RATIO |
| | Duration by Speed Variation | Duration × speed variation | $T \times CV_s$ | DUR_SOGCV |
| | Mean Speed by Length | Mean speed × length | $\bar{s} \times L$ | SOG_LEN |
| | Speed Standard Deviation by Draft | Speed std × draft | $\sigma_s \times D$ | SOGSTD_DFT |
| | Mean Speed by Width | Mean speed × width | $\bar{s} \times w$ | SOG_WID |
| | Square of Mean Speed | Square of mean speed | $\bar{s}^2$ | SOG_MEAN_SQ |
| | Square of Draft | Square of draft | $D^2$ | DFT_SQ |

where

$s = \{s_1, s_2, s_3, \cdots s_n\}$ is the series of n speed observations

$c = \{c_1, c_2, c_3, \cdots c_n\}$ is the series of n course observations

$A = \{a_1, a_2, a_3, \cdots a_{n-1}\}$ is the series of n acceleration values where $a_t = \frac{(s_{t+1} - s_t)}{\Delta t}$

L, W, D are the vessel's length, width and draft respectively

T is the total trip durationQ1 and Q3 are the first and third quartiles of the speed data.

$P(s_k)$ is the probability of speed falling into discrete bin k

## 3.2.    Machine Learning Models for Estimating Barge Tow Size from AIS Features

This section presents model selection and feature generation and selection.

*Model Selection*

Because barge count is a numeric outcome that takes non-negative integer values, regression-based models are appropriate. Regression models are designed to estimate continuous or count-based outcomes, making them ideal for forecasting the number of barges associated with each vessel based on vessel trajectory data. However, traditional regression models such as Ordinary Least Squares (OLS) assume linear, additive relationships among input features. These assumptions often break down when faced with high-dimensional, nonlinear, or interacting variables, as is common with AIS-derived features like speed entropy, course variability, and acceleration patterns. To address these limitations, this study employs Machine Learning (ML) regression models, which are capable of learning complex nonlinear functions, capturing interaction effects automatically, and generalizing better in heterogeneous real-world data environments.

Six diverse regression models were selected for comparison.  These represent a balance of flexibility, interpretability, and theoretical diversity: (1) Three ensemble-based models, (2) One kernel-based model, (3) One regularized linear model, and (4) One classical count-based model. By choosing models from different algorithmic families (e.g., ensemble, kernel-based, and linear) the study explores which techniques are best suited for this specific prediction task. The selected models are:

Ensemble Models

Three ensemble models were evaluated. These models combine the predictions of multiple base estimators to improve robustness and predictive accuracy. Ensembles are known for robustness and strong performance on structured data.  The *Random Forest Regressor* constructs a multitude of decision trees during training. For each tree, it uses a random subset of the data and a random subset of the features, a process which helps prevent overfitting and creates a more generalized model. The final prediction is the average of the predictions from all the individual trees, which is why it often achieves high accuracy (*41*).

The *Catboost Regressor* is a gradient-boosting algorithm optimized for performance. It uses a technique called "ordered boosting" to process data, which helps avoid bias and target leakage. While it's particularly known for its efficient handling of categorical features, its robust boosting mechanism makes it a strong contender for numerical datasets as well (*42*).  The *Adaboost Regressor* works by building a "strong" learner from a series of "weak" learners (typically simple decision trees). It iteratively places more weight on data points that were incorrectly predicted in previous rounds, forcing the model to focus on the most difficult cases. This adaptive process can lead to high accuracy, though it can sometimes be sensitive to noisy data (*43*).

<u>Kernel Based Model</u>

Kernel based models are well-suited for capturing high-dimensional nonlinear relationships. A *Support Vector Regressor (SVR)* was selected. Unlike other models that try to minimize error, SVR works by fitting a line (or hyperplane) that has the maximum number of data points within a specified margin or boundary. It uses the "kernel trick" to transform the input data into a higher-dimensional space, allowing it to model highly complex, non-linear relationships effectively (*44*).

<u>Regularized Linear Model</u>

*ElasticNet* was selected as a linear model. Regularized linear models provide strong baselines while handling collinearity. It is a hybrid of two other regularization methods: Ridge (L2) and Lasso (L1). The L2 component helps manage situations where predictor variables are highly correlated, while the L1 component performs automatic feature selection by shrinking the coefficients of less important features to zero. This combination makes it very effective for datasets with many, potentially collinear, features (*45*).

<u>Generalized Linear Model</u>

Generalized Linear models are commonly used for modeling discrete outcomes. A *Poisson Regressor* was selected. This model is specifically designed for regression tasks where the outcome is a count (a non-negative integer), like the number of barges. It's a generalized linear model that uses a logarithmic link function to ensure the predicted values are always non-negative. However, its performance can suffer if the data's variance is not equal to its mean (*46*).

*Feature Identification and Selection*

Features serve as predictors of barge count and are extracted from the AIS trajectories to describe:

- **Vessel Geometry:** Length, width, draft.
- **Speed Patterns:** Mean, variance, max/min, entropy.
- **Acceleration Behavior:** Magnitude, variability, directional changes.
- **Course Characteristics**: Course-over-ground statistics, turn rates, and heading deviations.
- **Trip Efficiency:** Travel duration, distance, sinuosity.
- **Interaction Terms:** Combinations of features (e.g., speed × length, draft²).

Recursive Feature Elimination with Cross-Validation (RFECV) (*47*) was employed for feature selection to identify the optimal subset of predictive features from over 30 initial variables. RFECV iteratively trains a model, ranks features by importance, and removes the least influential features while evaluating predictive performance via cross-validation. This procedure was applied separately across all six machine-learning models considered, tailoring feature selection to each model's learning characteristics. The final predictive models were then trained using only the selected feature subsets.

*Model Training Procedure*

To ensure robust model evaluation and reduce overfitting given the dataset (26 samples), a stratified K-fold cross-validation approach was used. Cross-validation repeatedly partitions the data into k folds, allowing the model to be trained on k–1 folds and validated on the remaining fold in each iteration. This repeated training and validation on different subsets provides more stable and reliable performance estimates than a single partition. The data was stratified by the number of barges attached to each vessel so that each fold preserved the overall distribution of barge count classes. This was important given the imbalanced and discrete nature of the outcome variable, as it prevented under- or over-representation of specific barge count classes in the training or validation sets. During the RFECV process, the model was iteratively trained and evaluated across all folds, with the negative Mean Absolute Error (MAE) serving as the performance metric to guide the selection of the optimal feature subset.

*Model Evaluation Metrics*

Each model was assessed using MAE (**Equation 4**). MAE is the absolute error in barge count estimation, i.e., how far off the prediction was from the actual count.

$$MAE = \frac{1}{n}\sum_{i=1}^{n}|x_i - \hat{x}_i| \hspace{3cm} \textbf{Equation 4}$$

where

  $n$ = The total number of vessel instances (e.g., tugs or towboats) in the test dataset for which barge counts are estimated.

  $x_i$ = The true number of barges attached to the i-th vessel, based on manual annotation or validated ground truth.

  $\hat{x}_i$ = The predicted number of barges attached to the i-th vessel, as estimated by the computer vision model.

## 4.   RESULTS AND DISCUSSION

A 240-mile section of the Mississippi River from Baton Rouge to the Gulf of Mexico served as a case study. This section is the nation's most active corridor for waterborne commercial traffic. Data was gathered between January and April 2024 (**Figure 3**). The following sections detail the data collection process, model development, and key findings.

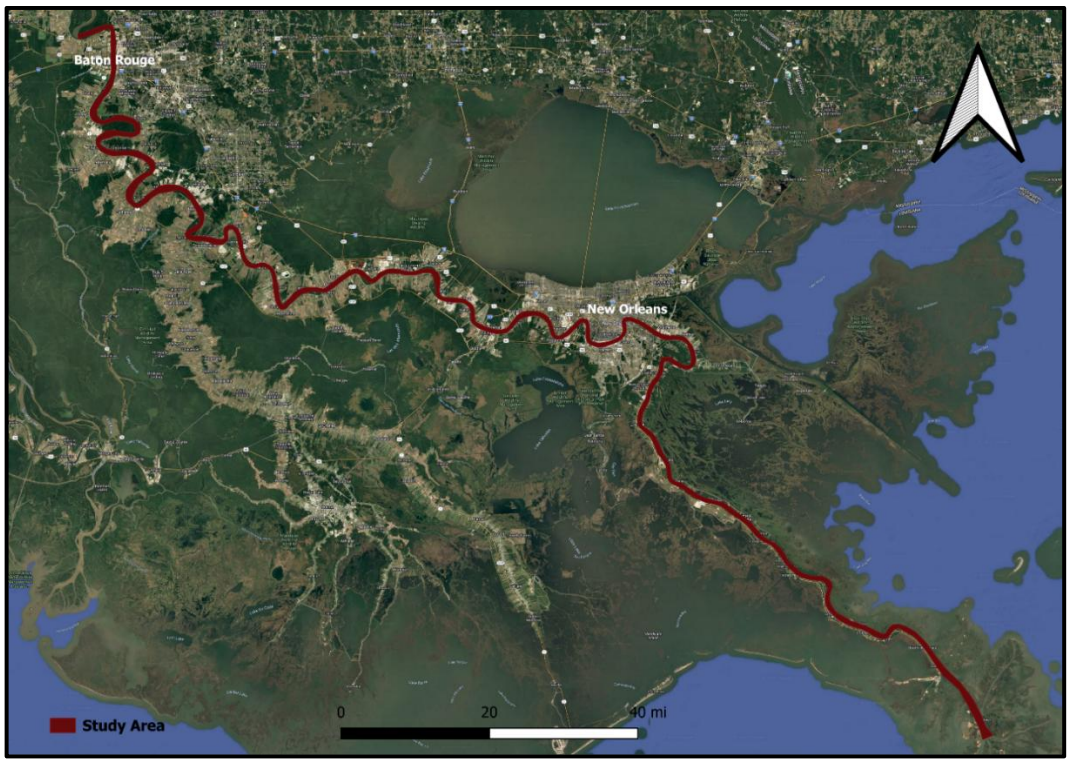

**Figure 3: Satellite map showing the study area along the Lower Mississippi River**

### 4.1. Data Acquisition and Image Annotation Results

High-resolution satellite imagery (3 m) from Planet Labs (*48*), updated daily with a 30-day delay, was used. A custom Application Programming Interface (API) was designed that processes AIS tracks for designated areas, detects their spatiotemporal overlap with Planet imagery tiles, and automatically downloads only the scenes containing visible vessels or barges. These targeted images served as labeled data for training our computer vision models. AIS data, recorded at 1-minute intervals across U.S. inland waterways, was sourced from the Marine Cadastre Web Tool (*49*).

Following the process outlined in Section 3.1, AIS points were matched to satellite images to generate manual sample labels. Then AIS trajectories were constructed after identifying the stop locations following Asborno et al (*40*). Parameters used in stop identification and subsequent trajectory construction include: a maximum speed threshold of 1.0 knot, minimum stop duration of 60 mins, and a spatial radius of 300m, selected through iterative testing and precision evaluation. Finally, from the AIS trajectories, 30 kinematic features were calculated following the formulations in **Table 1**.

The final annotated dataset consisted of 26 labeled instances, each with an associated barge count. These represent moving tows. Smaller barge configurations (e.g., two to four barges) were more frequently observed, while larger configurations (e.g., 12 or more barges) were less common (**Figure 4**). This distribution shows class imbalance which can influence prediction accuracy for less frequent barge counts.

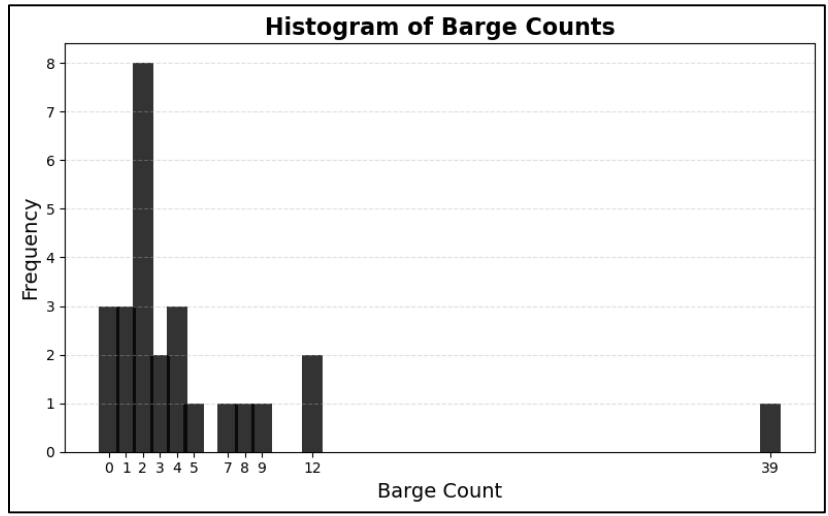

**Figure 4: Distribution of barge counts in the labeled dataset**

### 4.2. Barge Count Prediction via Machine Learning

Among the six regression models tested, the Poisson Regressor outperformed all others, achieving an average MAE of 1.920 barges across the two folds. This model used 12 of the 30 features. The Support Vector Regressor followed closely, with an MAE of 1.922 barges using one feature. AdaBoost ranked third, yielding an MAE of 2.784 barges with two features. The Random Forest Regressor produced an MAE of 3.525 barges based on four features. Finally, the Catboost Regressor and ElasticNet exhibited the highest error levels, with MAEs of 4.148 and 4.445 barges, respectively; Catboost employed 28 features, whereas ElasticNet used two. (**Table 2**)

**Table 2: Performance and Features of barge count prediction models**

| Model | MAE (barges) | # of Features | Features Used |
|---|---|---|---|
| **1. Poisson Regressor** | 1.920 | 12 | LEN, WID, SOG_CV, ACC_POS_MEAN, COG_STD, COG_ENT, TRN_STD, COG_HDG_DIFF_STD, SINO_IDX, DUR_SOGCV, SOG_LEN, SOG_WID |
| **2. Support Vector Regressor** | 1.922 | 1 | DUR_SOGCV |
| **3. AdaBoost** | 2.784 | 2 | COG_ENT, DIST_KM |
| **4. Random Forest Regressor** | 3.525 | 4 | ACC_ZC, COG_ENT, DIST_KM, DUR_SOGCV |
| **5. Catboost Regressor** | 4.148 | 28 | LEN, WID, SOG_MEAN, SOG_STD, SOG_MAX, SOG_MIN, SOG_CV, SOG_PCT_LOW, SOG_PCT_OPT, SOG_ENT, ACC_POS_MEAN, ACC_MIN, ACC_ZC, COG_STD, COG_ENT, COG_TOTAL_CHANGE, COG_HDG_DIFF_STD, DUR_HRS, DIST_KM, DIST_HAVERSINE_KM, SINO_IDX, AREA, DLT_RATIO, DUR_SOGCV, SOG_LEN, SOGSTD_DFT, SOG_WID, SOG_MEAN_SQ |
| **6. ElasticNet** | 4.445 | 2 | SOG_PCT_LOW, COG_STD |

A stratified 2-fold cross-validation scheme was employed for model evaluation. The data is stratified into two groups (folds) such that the first group is used to train the model and the second is used as an independent test set. The resulting test sets, referred to as folds 1 and 2 had 11 and 10 samples, respectively (**Figure 5**). A higher number of folds would have led to unreliable validation results, as certain folds would lack sufficient representation of rare tow sizes.

For the Poisson Regressor, in Fold 1 the model achieved an MAE of 2.482. It overpredicted some zero-barge cases (e.g. Observation 0: predicted 4.41 vs. actual 0) and underpredicted the single large tow (Observation 10: predicted 8.25 vs. actual 12), which drove up the fold's overall error. By contrast, Fold 2 produced a lower MAE of 1.301, showing that the model's predictions aligned more closely with most of the actual values. Aside from the extreme case in Observation 0 (predicted 6.00 vs. actual 12), most mid-range counts (Observations 1–9) were estimated within roughly ±1 barge of their true values. Together, these results indicate that while the Poisson Regressor is well-calibrated for the common, low-to-mid tow sizes, it still struggles with the rare, high-barge outliers.

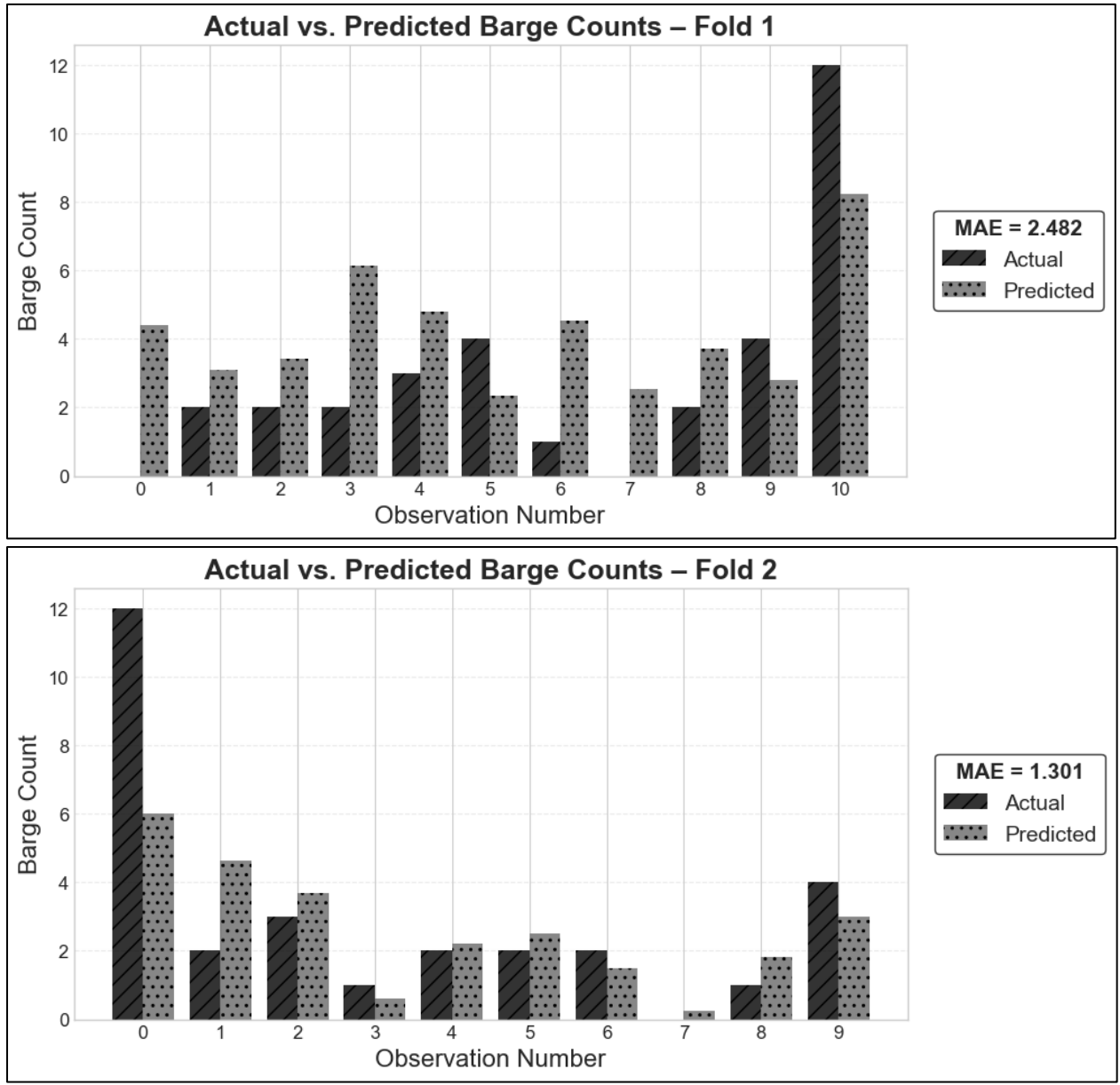

**Figure 5: Actual vs. predicted barge counts across two test folds using the Random Forest Regressor**

Entropy of Course Over Ground (COG_ENT) and Trip Duration interacted with Speed Variation (DUR_SOGCV) are jointly the most frequently selected features across all regression models (**Figure 6**). This aligns with the theoretical expectation that vessels towing larger barge configurations exhibit reduced maneuverability. COG_ENT is theoretically relevant because it quantifies the complexity of a vessel's heading profile. Each course change, whether a gentle bend or a sharp turn adds "information" to the trip's trajectory. Large barge tows, with their high inertia and limited turning radius, force tugboat operators to plan and execute only the most gradual, predictable maneuvers, smoothing out the heading signal. As a result, the sequence of headings carries far less variability and thus lower entropy than that of a nimble, lightly loaded vessel. DUR_SOGCV captures the interaction between trip duration and speed variability. Longer tows with many barges require very consistent speed profiles to safely manage increased mass and drag, so trips that combine extended durations with tightly regulated speeds become strong indicators of high barge counts.

Features not selected, such as maximum speed (SOG_MAX) or mean positive acceleration (ACC_POS_MEAN), were likely excluded due to redundancy, lower predictive power, or misalignment with behavioral patterns. For instance, SOG_MAX might reflect temporary conditions rather than tow size, while acceleration metrics may be incorporated by speed variability measures like SOG_CV. The RFECV process favored features that enhance generalization and reduce multicollinearity, prioritizing those like COG_ENT and DUR_SOGCV that capture the operational constraints imposed by barge count over an entire trip.

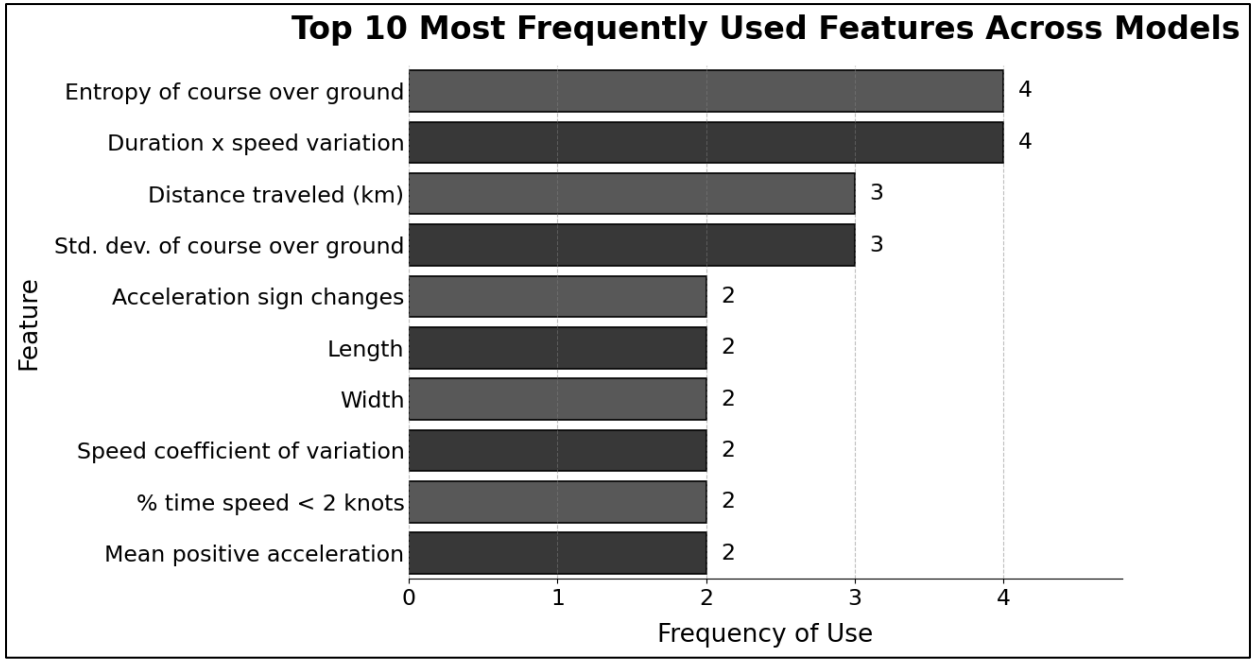

**Figure 6: Top 10 most frequently selected features across all regression models used to predict barge quantity**

# 5. CONCLUSION

This study demonstrates the feasibility of predicting barge quantity using AIS-derived vessel movement data, with satellite imagery employed solely to generate labeled training samples for model development. By leveraging widely available AIS data for predictions, the model can be applied in real-time scenarios without the need for continuous satellite coverage, making it practical and scalable for operational use.

Among six machine learning regressors evaluated, the Poisson Regressor model achieved the best performance with a MAE of 1.92 barges, using a subset of 12 (out of 30) features. Feature selection revealed that vessel maneuverability metrics, particularly 'course entropy' and the composite feature of 'trip duration multiplied by speed variation', were the two most influential predictors of barge count. The next two features, 'mean course change rate' and 'speed coefficient of variation' also capture aspects of directional control and propulsion stability that vary with tow size.

Future work aims to enhance this feasibility study in several ways. First, the model training and evaluation were constrained by sample size. Expanding the dataset to include a broader range of vessel types, tow configurations, and geographies, will improve model generalizability. Second, the modeling framework does not explicitly account for environmental conditions that can significantly affect vessel behavior and thus, indirectly, barge count predictions. These include river flow velocity and direction, wind speed and direction, water levels, turbulence, visibility conditions (e.g., fog, rain), channel width, and seasonal hydrologic variability. While some geometric characteristics of the river, such as sinuosity, were encoded in the feature set, real-time or historical hydrodynamic data were not incorporated for this proof-of-concept. The absence of these contextual variables may partially explain residual errors in predictions. Third, while this study evaluated a diverse set of regression models, including ensemble-based, kernel-based, and generalized linear approaches, more advanced deep learning architectures such as recurrent neural networks (RNNs) or graph neural networks (GNNs) may enhance performance by capturing complex spatial and temporal dependencies. Finally, future work will incorporate a spatial generalization framework to evaluate model performance across rivers with differing characteristics, such as the Ohio River (wider channel, higher traffic density), the Arkansas River (more constrained navigation), and the Columbia River (influenced by tidal patterns and hydropower infrastructure).

## 6. ACKNOWLEDGEMENTS

The authors acknowledge the support and sponsorship provided by the National Science Foundation (NSF) and the U.S. Army Corps of Engineers (USACE). The authors acknowledge the use of OpenAI's ChatGPT to assist with language refinement. All analyses, results, and interpretations were conducted independently by the authors.

## 7. AUTHOR CONTRIBUTIONS

The authors confirm their contributions to the paper as follows: study conception and design: G. Agorku, S. Hernandez; data collection: G. Agorku, H. Hames and C. Wagner; analysis and interpretation of results: G. Agorku, S. Hernandez; draft manuscript preparation: G. Agorku and S. Hernandez.

## 8. DECLARATION OF CONFLICTING INTERESTS

The authors declare that they have no known competing financial interests or personal relationships that could have appeared to influence the work reported in this paper.

## 9. FUNIDNG

This research was supported by the National Science Foundation (NSF) [Award number 2042870].

## 10. REFERENCES

1. Full-Report-2025-Natl-IRC-WEB.Pdf. .
2. Agorku, G., S. V. Hernandez, M. Falquez, S. Poddar, and K. Amankwah-Nkyi. Real-Time Barge Detection Using Traffic Cameras and Deep Learning on Inland Waterways. *Transportation Research Record: Journal of the Transportation Research Board*, 2024.
3. Accessibility_and_Usability_of_AIS_Data.Pdf. .
4. Dansarie, M. Security Issues in Special-Purpose Digital Radio Communication Systems: A Systematic Review. *IEEE Access*, Vol. 12, 2024, pp. 91101–91126. https://doi.org/10.1109/ACCESS.2024.3420091.
5. Niu, T., Z. Shao, and G. Zhu. Toward Greener Freight: Overview of Inland Waterway Transport for Freight in the European Union.
6. Gobert, J., and F. Rudolf. Rhine Low Water Crisis: From Individual Adaptation Possibilities to Strategic Pathways. *Frontiers in Climate*, Vol. 4, 2023. https://doi.org/10.3389/fclim.2022.1045446.
7. Bcg-Freight-Rails-Digital-Future-Is-Just-around-the-Bend-Jan-2023-r.Pdf. .
8. (PDF) Impacts of a Tracking and Tracing System for Containers in a Port-Based Supply Chain. *ResearchGate*. https://doi.org/10.14488/BJOPM.2016.v13.n3.a12.
9. Gordebeke, G.-J., S. Lemey, O. Caytan, M. Boes, J. Jocqué, S. Van de Velde, C. Marshall, E. De Poorter, and H. Rogier. Time-Domain-Optimized Antenna Array for High-Precision IR-UWB Localization in Harsh Urban Shipping Environments. *IEEE Sensors Journal*, Vol. 24, No. 5, 2024, pp. 5561–5577. https://doi.org/10.1109/JSEN.2023.3310992.
10. Dobrkovic, A., M.-E. Iacob, and J. van Hillegersberg. Maritime Pattern Extraction and Route Reconstruction from Incomplete AIS Data. *International Journal of Data Science and Analytics*, Vol. 5, No. 2, 2018, pp. 111–136. https://doi.org/10.1007/s41060-017-0092-8.
11. Asborno, M. I., and S. Hernandez. Assigning a Commodity Dimension to AIS Data: Disaggregated Freight Flow on an Inland Waterway Network. *Research in Transportation Business & Management*, Vol. 44, 2022, p. 100683. https://doi.org/10.1016/j.rtbm.2021.100683.
12. Erlbaum, N., and J. Holguin-Veras. Some Suggestions for Improving CFS Data Products. Presented at the Commodity Flow Survey ConferenceFederal Highway AdministrationU.S. Census BureauAmerican Association of State Highway and Transportation Officials (AASHTO)Research and Innovative Technology AdministrationTransportation Research Board, 2006.
13. Kam, K. A., N. Jiang, P. Bujanovic, K. M. Savage, R. Walthall, D. Seedah, and C. M. Walton. Finding and Exploring Use of Commodity-Specific Data Sources for Commodity Flow Modeling. *Transportation Research Record*, Vol. 2646, No. 1, 2017, pp. 77–83. https://doi.org/10.3141/2646-09.
14. Bhurtyal, S., H. Bui, S. Hernandez, S. Eksioglu, M. Asborno, K. N. Mitchell, and M. Kress. Prediction of Waterborne Freight Activity with Automatic Identification System Using Machine Learning. *Computers & Industrial Engineering*, Vol. 200, 2025, p. 110757. https://doi.org/10.1016/j.cie.2024.110757.
15. (PDF) Design of a Web Based Information System to Manage Delivery Activities. *ResearchGate*. https://doi.org/10.25124/jrsi.v10i02.721.
16. Ruan, S., X. Fu, C. Long, Z. Xiong, J. Bao, R. Li, Y. Chen, S. Wu, and Y. Zheng. Filling Delivery Time Automatically Based on Couriers' Trajectories. *IEEE Transactions on Knowledge and Data Engineering*, Vol. 35, No. 2, 2023, pp. 1528–1540. https://doi.org/10.1109/TKDE.2021.3100116.
17. Agorkua, G., S. Hernandez, M. Falquez, S. Poddar, and S. Pang. Predicting Barge Presence and Quantity on Inland Waterways Using Vessel Tracking Data: A Machine Learning Approach. http://arxiv.org/abs/2501.00615. Accessed Jul. 11, 2025.
18. Lee, J., E. Lin, M. Wang, and S. Maity. Computer Vision for Detection of Illegal Mining Barges in the Rio Madeira.

19. Power of ERDC Podcast Ep. #25: CorpsCam: Enabling More Proactive Coastal Management through Real-Time Monitoring and Data. https://www.erdc.usace.army.mil/Media/Video-Page/audioid/75569/. Accessed Jul. 30, 2025.

20. Ling, Y., M. Jin, M. R. Hilliard, and J. M. Usher. A Study of Real-Time Identification and Monitoring of Barge-Carried Hazardous Commodities. Presented at the 2009 17th International Conference on Geoinformatics, 2009.

21. Sakib, S., and M. S. Bin Abdullah. GPS-GSM Based Inland Vessel Tracking System for Automatic Emergency Detection and Position Notification. Presented at the 2016 10th International Conference on Intelligent Systems and Control (ISCO), 2016.

22. Ophoff, T., S. Puttemans, V. Kalogirou, J.-P. Robin, and T. Goedemé. Vehicle and Vessel Detection on Satellite Imagery: A Comparative Study on Single-Shot Detectors. *Remote Sensing*, Vol. 12, No. 7, 2020, p. 1217. https://doi.org/10.3390/rs12071217.

23. Farr, A. J., I. Petrunin, G. Kakareko, and J. Cappaert. Self-Supervised Vessel Detection from Low Resolution Satellite Imagery. In *AIAA SCITECH 2022 Forum*, American Institute of Aeronautics and Astronautics.

24. Sankhe, D., and S. Bhosale. Vessel Detection in Satellite Images Using Deep Learning. *Engineering, Technology & Applied Science Research*, Vol. 14, No. 6, 2024, pp. 18357–18362. https://doi.org/10.48084/etasr.8755.

25. Jesawada, A., H. Singh, K. Patel, D. Kumari, and R. Bhosale. AI-Driven Marine Vessel Detection Through Satellite Imagery: A Deep Learning Approach. *IJFMR - International Journal For Multidisciplinary Research*, Vol. 6, No. 6, 2024. https://doi.org/10.36948/ijfmr.2024.v06i06.34231.

26. Karantaidis, I., K. Bereta, and D. Zissis. A Hybrid Method for Vessel Detection in High-Resolution Satellite Imagery. Presented at the IGARSS 2023 - 2023 IEEE International Geoscience and Remote Sensing Symposium, 2023.

27. Bakirci, M. Advanced Ship Detection and Ocean Monitoring with Satellite Imagery and Deep Learning for Marine Science Applications. *Regional Studies in Marine Science*, Vol. 81, 2025, p. 103975. https://doi.org/10.1016/j.rsma.2024.103975.

28. Hall, D., and J. Llinas, Eds. *Multisensor Data Fusion*. CRC Press, Boca Raton, 2001.

29. Mazzarella, F., A. Alessandrini, W. G. H. Van, A. M. Alvarez, P. Argentieri, D. Nappo, and L. W. Ziemba. Data Fusion for Wide-Area Maritime Surveillance. *JRC Publications Repository*. https://publications.jrc.ec.europa.eu/repository/handle/JRC81422. Accessed Jul. 29, 2025.

30. Kanjir, U., H. Greidanus, and K. Oštir. Vessel Detection and Classification from Spaceborne Optical Images: A Literature Survey. *Remote Sensing of Environment*, Vol. 207, 2018, pp. 1–26. https://doi.org/10.1016/j.rse.2017.12.033.

31. Yu, T. E., B. P. Sharma, and B. C. English. Investigating Lock Delay on the Upper Mississippi River: A Spatial Panel Analysis. *Networks and Spatial Economics*, Vol. 19, No. 1, 2019, pp. 275–291. https://doi.org/10.1007/s11067-018-9395-0.

32. Foltz, S. *Investigation of Mechanical Breakdowns Leading to Lock Closures*. Construction Engineering Research Laboratory (U.S.), 2017.

33. Davis, J. P. Problems of Inland Waterway Lock Dimensions. *Journal of the Waterways, Harbors and Coastal Engineering Division*, Vol. 96, No. 2, 1970, pp. 451–466. https://doi.org/10.1061/AWHCAR.0000028.

34. Pazan, K. I., N. P. Memarsadeghi, J. C. Hodges, and C. and H. Laboratory (U.S.). Lock and Dam 25, Upper Mississippi River Navigation Study : Ship-Simulation Results. 2024.

35. Hammedi, W., S. M. Senouci, P. Brunet, and M. Ramirez-Martinez. Two-Level Optimization to Reduce Waiting Time at Locks in Inland Waterway Transportation. *ACM Trans. Intell. Syst. Technol.*, Vol. 13, No. 6, 2022, p. 91:1-91:30. https://doi.org/10.1145/3527822.

36. Guan, H., Y. Xu, L. Li, and X. Huang. Optimizing Lock Operations and Ship Arrivals through Multiple Locks on Inland Waterways. *Mathematical Problems in Engineering*, Vol. 2021, No. 1, 2021, p. 6220559. https://doi.org/10.1155/2021/6220559.

37. Hrušovský, M., E. Demir, W. Jammernegg, and T. Van Woensel. Real-Time Disruption Management Approach for Intermodal Freight Transportation. *Journal of Cleaner Production*, Vol. 280, 2021, p. 124826. https://doi.org/10.1016/j.jclepro.2020.124826.

38. Milne, D., and D. Watling. Big Data and Understanding Change in the Context of Planning Transport Systems. *Journal of Transport Geography*, Vol. 76, 2019, pp. 235–244. https://doi.org/10.1016/j.jtrangeo.2017.11.004.

39. Jocher, G., A. Chaurasia, and J. Qiu. YOLO by Ultralytics. Jan, 2023.

40. Asborno, M. I., S. Hernandez, K. N. Mitchell, and M. Yves. Inland Waterway Network Mapping of AIS Data for Freight Transportation Planning. *The Journal of Navigation*, Vol. 75, No. 2, 2022, pp. 251–272. https://doi.org/10.1017/S0373463321000953.

41. Breiman, L. Random Forests. *Machine Learning*, Vol. 45, No. 1, 2001, pp. 5–32. https://doi.org/10.1023/A:1010933404324.

42. Prokhorenkova, L., G. Gusev, A. Vorobev, A. V. Dorogush, and A. Gulin. CatBoost: Unbiased Boosting with Categorical Features. No. 31, 2018.

43. Freund, Y., and R. E. Schapire. A Desicion-Theoretic Generalization of on-Line Learning and an Application to Boosting. Berlin, Heidelberg, 1995.

44. Smola, A. J., and B. Schölkopf. A Tutorial on Support Vector Regression. *Statistics and Computing*, Vol. 14, No. 3, 2004, pp. 199–222. https://doi.org/10.1023/B:STCO.0000035301.49549.88.

45. Zou, H., and T. Hastie. Regularization and Variable Selection Via the Elastic Net. *Journal of the Royal Statistical Society Series B: Statistical Methodology*, Vol. 67, No. 2, 2005, pp. 301–320. https://doi.org/10.1111/j.1467-9868.2005.00503.x.

46. McCullagh, P. *Generalized Linear Models*. Routledge, New York, 2019.

47. Guyon, I., J. Weston, S. Barnhill, and V. Vapnik. Gene Selection for Cancer Classification Using Support Vector Machines. *Machine Learning*, Vol. 46, No. 1, 2002, pp. 389–422. https://doi.org/10.1023/A:1012487302797.

48. Planet Satellite Imaging | Planet. https://www.planet.com/. Accessed Jun. 22, 2025.

49. MarineCadastre.Gov | Vessel Traffic Data. https://marinecadastre.gov/ais/. Accessed Aug. 31, 2023.